# Fractional Wavelet Scattering Network and Applications

Li Liu, Jiasong Wu, *Member, IEEE*, Dengwang Li, Lotfi Senhadji, *Senior Member, IEEE*, and Huazhong Shu, *Senior Member, IEEE*

*Abstract—Objective:* The present study introduces a fractional wavelet scattering network (FrScatNet), which is a generalized translation invariant version of the classical wavelet scattering network (ScatNet). *Methods:* In our approach, the FrScatNet is constructed based on the fractional wavelet transform (FRWT). The fractional scattering coefficients are iteratively computed using FRWTs and modulus operators. The feature vectors constructed by fractional scattering coefficients are usually used for signal classification. In this work, an application example of FrScatNet is provided in order to assess its performance on pathological images. Firstly, the FrScatNet extracts feature vectors from patches of the original histological images under different orders. Then we classify those patches into target (benign or malignant) and background groups. And the FrScatNet property is analyzed by comparing error rates computed from different fractional orders respectively. Based on the above pathological image classification, a gland segmentation algorithm is proposed by combining the boundary information and the gland location. *Results:* The error rates for different fractional orders of FrScatNet are examined and show that the classification accuracy is significantly improved in fractional scattering domain. We also compare the FrScatNet based gland segmentation method with those proposed in the 2015 MICCAI Gland Segmentation Challenge and our method achieves comparable results. *Conclusion:* The FrScatNet is shown to achieve accurate and robust results. More stable and discriminative fractional scattering coefficients are obtained by the FrScatNet in this work. *Significance:* The added fractional order parameter is able to analyze the image in the fractional scattering domain.

*Index Terms*—fractional wavelet transform (FRWT), scattering network, classification, histopathology image, gland segmentation.

This work was supported in part by the National Key R&D Program of China (2017YFC107900) and in part by the National Natural Science Foundation of China (Nos. 61201344, 61271312, 61401085, 31571001, 31640028, 31400842, 61572258, 11301074), by the Qing Lan Project and the '333' project (No. BRA 2015288), and by the Short-term Recruitment Program of Foreign Experts (WQ20163200398).

Li Liu, Jiasong Wu, and Huazhong Shu are with the Laboratory of Image Science and Technology, the Key Laboratory of Computer Network and Information Integration (Southeast University), Ministry of Education, 210096 Nanjing, China, and with the International Joint Laboratory of Information Display and Visualization, 210096 Nanjing, China, and also with the Centre de Recherche en Information Biomédicale Sino-Français (CRIBs), 210096 Nanjing, China (e-mail: liuli19910101@163.com, jswu@seu.edu.cn, shu.list@seu.edu.cn).

Lotfi Senhadji is with INSERM, U1099, 35000 Rennes, France, the Laboratoire Traitement du Signal et de l'Image (LTSI), Université de Rennes 1, 35000 Rennes, France, and the Centre de Recherche en Information Biomédicale Sino–Français (CRIBs), 35000 Rennes, France (e-mail: lotfi.senhadji@univ-rennes1.fr).

Dengwang Li is with Shandong Province Key Laboratory of Medical Physics and Image Processing Technology, School of Physics and Electronics, Shandong Normal University, 250014 Jinan, China (e-mail : dengwang@sdnu.edu.cn).

## I. INTRODUCTION

IN most cases, when dealing with images, the conveyed information varies considerably over time and space. Extracting robust features and measuring similarities in order to get an effective image analysis is not straightforward [1]. The presence of noise, deformations and non-stationary behavior make it difficult to segment [2, 3] or to classify [4] signals. In image processing, textural pattern, as one of robust features, is usually used to describe surface object properties and their relationships to the surrounding environment [5].

The Wavelet Transform (WT) is one of the most widely used tools for transient and non-stationary signals and image texture analysis. It is able to simultaneously describe signals in both time and frequency domains although some limitations have been pointed out [6]. To overcome these difficulties, fractional Fourier transform (FRFT) has been proposed [7], generalizing the Fourier transform (FT) and extending the time-frequency plane to the time-fractional-frequency plane [8]. However, the fractional Fourier representation fails in locating the occurrence of the FRFT spectral content at a particular time due to its global kernel [9], which is indispensable when analyzing non-stationary signals. A modification, called short time FRFT [10], is developed to try to overcome this time location problem. But the technique is limited by the fundamental uncertainty principle in its application [11]. Like the FRFT, the fractional WT (FRWT) derived in [12] is aimed at representing the fractional spectrum. Constructed on the FRFT, it also fails in obtaining the local information of the signal. Many other FRWTs are also developed to jointly display time and FRFD (fractional Fourier domain)-frequency information of a signal [13-15]. More general FRWT methods are proposed based on fractional convolution [16, 17]. The FRWT reported in [16] is easy to implement and has low computational complexity. The FRWT described in [17] depicts more mathematical properties. In this study, we choose the FRWT presented in [16] to construct the fractional wavelet scattering network (FrScatNet).

For classification purpose, a feature vector must be set. Wavelet coefficients are often used. Because WTs are not invariant to translation, a scattering operator that is invariant to translation and rotation is built from the wavelet coefficients [18]. The translation invariant representation is computed by cascading WTs and modulus pooling operators, which averages the amplitude of iterated wavelet coefficients [1, 18, 19]. Due to the WT limitations in the time-frequency domain, if the energy concentration of a signal is not optimal in the frequency domain, the wavelet coefficients are not the best representation of its energy distribution. So, the detected scattering coefficients are not the most discriminative features for signal classification. In this study, the fractional scattering coefficients are computed by cascading FRWT and averaging the amplitude of iterated fractional wavelet coefficients in order to address this issue. The obtained FrScatNet generalizes the classical wavelet scattering network (ScatNet) from the scattering domain to the fractional scattering domain. The added fractional order parameter is then able to analyze the signal in the fractional scattering domain. We test the efficiency of the FrScatNet on two-dimensional images in this study. Besides, we also use it to achieve the gland segmentation in colon histology images of tissue slides stained with Hematoxylin and Eosin (H&E).

Let us briefly summarize the state-of-the-art in this very challenging research area. Histopathological biopsy evaluation is considered as the gold standard for colon, prostate, and breast cancer diagnosis, malignancy confirmation and grading [20, 21]. Colorectal adenocarcinoma is the most common form of colon cancer. It is of fundamental importance to achieve good intra-observer and inter-observer reproducibility in the cancer grading. An automated method with the capability to detect the morphological information can be used as computer assistant for cancer grading. There are plenty of difficulties existing in quantifying the morphology of gland automatically. The structure, morphology, size and location of a gland vary significantly. Especially, cancer progresses may cause changes in the component organization, and also lead to tissue degeneration. In addition, some glands are even touching to each other leading to the coalescence problem [22]. Various approaches have been proposed to achieve gland segmentation, such as texture based methods [23-25], and structure based methods [26-30]. The drawbacks of all these methods are that they solely use pixel-level color information while assuming a regular architecture of glandular structures. Recently, convolutional neural networks have been considered. Kainz et al [31] combined deep convolutional neural network and total variation. Li et al [32] used handcrafted features and convolutional neural networks to recognize glands. Xu et al [33] presented a multichannel learning approach to extract region, boundary and location cues. In the MICCAI 2015 gland segmentation challenge contest [28], the convolutional neural networks were used and achieved impressive performance. Chen et al [34] presented a novel deep contour-aware network. Sirinukunwattana et al [35] proposed a multipath convolutional neural network segmentation algorithm. Ronneberger et al [36] applied a u-shaped deep convolutional network. Despite the good results obtained by these neural networks, the involved cascaded nonlinearities make their properties and optimal configurations not clear. The scattering networks address these questions from a mathematical and algorithmic perspective by concentrating on a particular class of deep convolutional networks [1]. Because the scattering network is learning-free due to the fixed wavelet basis, it can be implemented easily with less parameters and hardware resource.

In this paper, we propose a generalized FrScatNet based on the FRWT and use it to classify signals and solve the challenging problem in gland segmentation. Section II describes the data sets onto which our experiments are conducted and our proposed fractional scattering network. Section III illustrates the results achieved so far and compares our approach to other reference methods. A discussion is also provided before concluding (Section IV).

## II. METHODS

The used public data sets are firstly sketched and then a full presentation of the FrScatNet is given.

### A. Data sets

The medical dataset Warwick-QU is provided by the Colon Histology Images Challenge Contest held at MICCAI'2015 [35, 37]. The dataset consists of 165 images derived from 16 H&E stained colon histological slides of stage T3 or T42 colorectal adenocarcinoma. The slides are digitally scanned at 20 magnifications by Zeiss MIRAX MIDI Slide Scanner with a pixel resolution of 0.465μm. The whole-slide images are subsequently rescaled to a pixel resolution of 0.620μm. Each section belongs to a different patient. The dataset is grouped into 85 training images, 60 testing images (test A) and 20 testing images (test B). The annotations are regarded as ground truth. The dataset exhibits high inter-subject variability in both stain distribution and tissue architecture.

The public texture database created in [38] includes surfaces with significant viewpoint changes and scale differences within each class, whose texture is due mainly to albedo variations, 3D shapes, and a mixture of both. The database is also chosen as the benchmark for texture recognition. It contains 25 texture classes and each class has 40 examples. The resolution of the samples is 640×480 pixels.

### B. The Proposed Fractional Scattering Network

In this section, the formulation of the FrScatNet is described. We start by introducing the definition of the FRWT. Then, a FRWT-based scattering network is proposed. In the following, the convolution network structure is described.

#### 1) Definition of the FRWT

According to the definition in [9, 39], the FRFT can provide signal representation in the fractional Fourier domain, but it fails in obtaining the local structures of the signal. FRWT is able to handle the problem and display signal information in the fractional wavelet domain.

For the continuous-time finite energy signal $x(t)$ (i.e. signal belonging to $L^2(\mathbb{R})$), the WT is defined as follows:

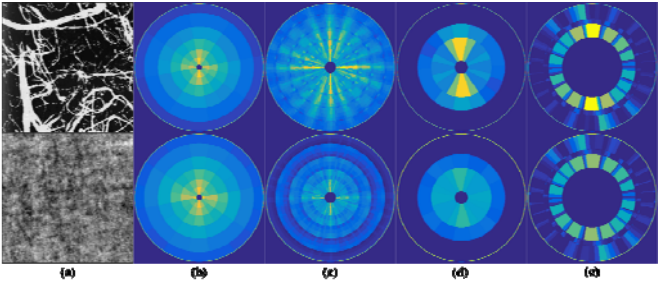

Fig. 1. (a) Realizations of two stationary processes $X(\mu)$. Top: Brodatz texture. Bottom: Gaussian process. (b) First-order scattering coefficients $S[\lambda_1]x$ from the ScatNet [1] are nearly the same ($\alpha_1$=1.00, $\alpha_2$=1.00). (c) Second-order scattering coefficients $S[\lambda_1, \lambda_2]x$ from the ScatNet [1] are clearly different ($\alpha_1$=1.00, $\alpha_2$=1.00). (d) First-order fractional scattering coefficients $S[\lambda_1]x$ from the FrScatNet are clearly different ($\alpha_1$=1.00, $\alpha_2$=0.70 ).(e) Second-order fractional scattering coefficients $S[\lambda_1, \lambda_2]x$ from FrScatNet are also clearly different ($\alpha_1$=1.00, $\alpha_2$=0.70 ).

$$W_x(a,b) = \frac{1}{\sqrt{a}} \int_{\mathbb{R}} x(t)\psi^*\left(\frac{t-b}{a}\right) dt, \quad (1)$$

where $\psi(t)$ is the mother wavelet. The superscript * denotes complex conjugate. The compressed or dilated wavelet $\psi_{a,b}(t) = (1/a^{1/2})\cdot\psi((t-b)/a)$ is obtained by the affine transformation of the mother wavelet. $a \in \mathbb{R}^+$ is the continuous scale parameter and $b \in \mathbb{R}$ represents the shift of the mother wavelet along the $t$ domain. The different scale WT intrinsically behaves as frequency band-pass filters, which describes the signal in the time-frequency plane. FRWT has been proposed as an extension of WT to analyze the time-varying FRFT spectra [13, 17, 40].

Different definitions of fractional convolution have been proposed in the literature. According to the definition introduced in [41], the fractional convolution of two signals $x(t)$ and $h(t) \in L^2(\mathbb{R})$ is expressed as follows:

$$x(t)\Theta_\alpha h(t) = \int_{\mathbb{R}} x(\tau)h(t-\tau)e^{-j\frac{t^2-\tau^2}{2}\cot\theta} d\tau \quad (2)$$
$$= e^{-\frac{j}{2}t^2\cot\theta}[(x(t)e^{\frac{j}{2}t^2\cot\theta}) * h(t)],$$

where $\Theta_\alpha$ denotes the fractional convolution operator. The parameter $\alpha$ is the fractional order and $\theta = \alpha\pi/2$ represents the rotation angle.

A kind of FRWT with simple structure and easy implementation is proposed in [13, 16]. It is constructed based on the fractional convolution according to the relationship between WT and the classical convolution shown in equation (1). For a given signal $x(t) \in L^2(\mathbb{R})$, the $\alpha$ order FRWT is:

$$W_x^\alpha(a,b) = x(t)\Theta_\alpha(\frac{1}{\sqrt{a}}\psi^*(-\frac{t}{a}))$$
$$= \int_{\mathbb{R}} x(t)\psi_{\alpha,a,b}^*(t)dt \quad (3)$$
$$= e^{-(j/2)b^2\cot\theta}\int_{\mathbb{R}} (x(t)e^{(j/2)t^2\cot\theta})\psi_{a,b}^*(t)dt,$$

where the kernel function is defined by multiplying the classical wavelet $\psi_{a,b}(t)$ with the following chirp signal :

$$\psi_{\alpha,a,b}(t) = \psi_{a,b}(t)e^{-(j/2)(t^2-b^2)\cot\theta}. \quad (4)$$

We only consider the parameter $0 < \theta < \pi$, and it is easy to extend outside the interval $[0, \pi]$. When $\alpha = 1$, the FRWT defined in equation (3) reduces to conventional WT. The FRWT on different scales also behaves as a set of band-pass filters.

*2) Fractional Scattering Wavelets*

In this step, a scattering transform computes nonlinear invariants from fractional wavelet coefficients by modulus operator chosen as a nonlinear pooling operator since it has the capability to preserve the signal energy. For the complex signal $x(t) = x_r(t) + jx_i(t)$ ($x_r(t), x_i(t) \in L^2(\mathbb{R})$), the modulus operator is defined as $|x(t)|=(|x_r(t)|^2+|x_i(t)|^2)^{1/2}$. The wavelet-modulus coefficients, termed as the translation invariant coefficients, are built from the fractional wavelet by the modulus operator defined as:

$$U[\lambda]x = |x(t)\Theta_\alpha\psi_\lambda|. \quad (5)$$

More fractional wavelet-modulus coefficients can be obtained by further iterating on the FRWTs and the modulus operator along any path. The scattering propagator $U[p]$ for a given signal $x(t) \in L^2(\mathbb{R})$ is defined by cascading fractional wavelet-modulus operators:

$$U[p]x = U[\lambda_m]\cdots U[\lambda_2]U[\lambda_1]x$$
$$= |||x\Theta_\alpha\psi_{\lambda_1}|\Theta_\alpha\psi_{\lambda_2}|\cdots|\Theta_\alpha\psi_{\lambda_m}|, \quad (6)$$

where $U[\emptyset]x = x$ and $\emptyset$ denotes an empty set. A path is defined by the sequence $p = (\lambda_1, \lambda_2, \ldots, \lambda_m)$ with a length of $m$.

For classification purpose, the local descriptors are computed by defining a windowed fractional scattering transform with a scaled spatial window $\phi_{2^J}$:

$$S[p]x = U[p]x\Theta_\alpha\phi_{2^J}$$
$$= |||x\Theta_\alpha\psi_{\lambda_1}|\Theta_\alpha\psi_{\lambda_2}|\cdots|\Theta_\alpha\psi_{\lambda_m}|\Theta_\alpha\phi_{2^J}, \quad (7)$$

where $S[\emptyset]x = x\Theta_\alpha\phi_{2^J}$. The fractional scattering operator $S$ performs a spatial averaging on a domain whose width is proportional to $2^J$. The resulting windowed scattering is nearly invariant to a translation.

*3) Fractional Scattering Network*

For the path $p = (\lambda_1, \lambda_2, \ldots, \lambda_m)$, a convolution network is constructed by iterating the fractional scattering propagator $\hat{W}$, which is built based on the fractional scattering transform to compute the fractional complex wavelet coefficient modulus and filter the lower frequency. In this section a convolution network with three layers is constructed by iterating the fractional scattering propagator $\hat{W}$ (Fig. 2). We generalize the classical WT to FRWT, so the fractional wavelet coefficients are obtained from different fractional wavelet domain. The built FrScatNet has the ability to analyze the image in fractional scattering domain. The fractional scattering propagator $\hat{W}$ is applied to the input signal $x$ to compute the first layer of fractional wavelet coefficient modulus $U[\lambda_1]x$ and output its local average $S[\varnothing]x$. Applying $\hat{W}$ to all propagated signals $U[p]x$ of the $m$th layer outputs fractional scattering signals $S[p]x$ and computes all propagated signals on the next layer.

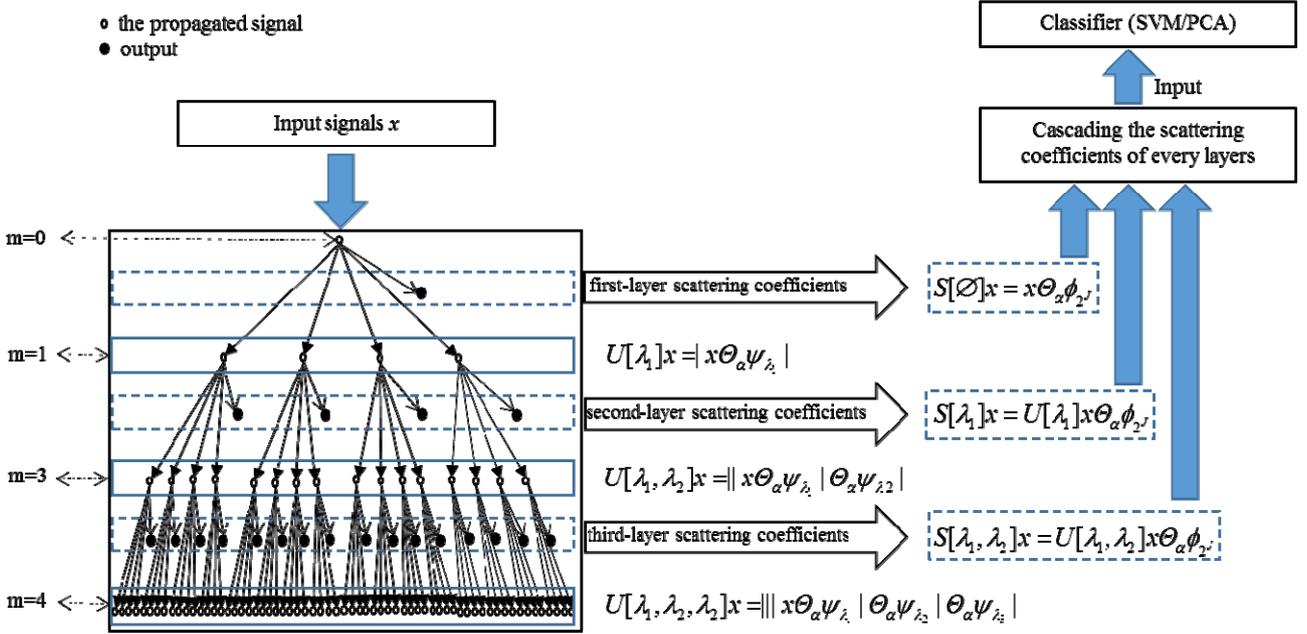

Fig. 2. An illustration of classification application of the FrScatNet.

The scattering transform is non-expensive and preserves the signal norm by given conditions like in work [1]. According to the Proposition 2.1 in [18], if the input signal $f$ is complex and the following rule is obeyed:

$$\forall J \in \mathbb{Z}, |\hat{\phi}(2^J \omega)|^2 + \sum_{j<J, r \in G} |\hat{\psi}(2^j r \omega)|^2 = 1 \quad (8)$$

Then $\|W_f\|^2 = \|f\|^2$ is obtained by using Plancherel formula. The similar condition can be given. If there exists $\varepsilon > 0$, such that for all $\omega \in \mathbb{R}^2$,

$$1 - \varepsilon \leq |\hat{\phi}(\omega)|^2 + \sum_{j=0}^{\infty} \sum_{r \in G} |\hat{\psi}(2^j r \omega)|^2 \leq 1 \quad (9)$$

Applying the Plancherel formula, we can deduce that if $f$ is complex, then $W_f$ satisfies $(1-\varepsilon)\|f\|^2 \leq \|W_f\|^2 \leq \|f\|^2$ with $\|W_f\|^2 = \|f * \phi_{2^J}\|^2 + \sum_{r \in G} \|f * \psi_\lambda\|^2$ [18]. According to the fractional wavelet transform in equation (2). The norm of $Wx(u) = \{x \Theta_\alpha \phi_{2^J}(u), x \Theta_\alpha \psi_\lambda(u)\}_{\lambda \in P}$ ($P = \{\lambda = 2^{-j}r : r \in G, j \leq J\}$) satisfies $\|W\|^2 = \|W_f\|^2$ with $f = x(u)e^{jt^2 \cot\theta/2}$. $f$ is a complex function and $\|f\|^2 = \|x\|^2$. So we can get that $(1-\varepsilon)\|x\|^2 \leq \|W\|^2 \leq \|x\|^2$. We suppose that $\varepsilon < 1$ and the fractional wavelet transform $W$ is non-expansive. If $\varepsilon = 0$, then $W$ is unitary and it is able to contain the norm of $x$. In this work, we use two kinds of wavelet transform to construct FrScatNet. For the Morlet wavelet, the value of parameter $\sigma$ in $\phi$ is set as 0.7. The value of $\sigma$ in $\psi$ is 0.5. And equation (10) is satisfied with $\epsilon = 0.98$. $\phi$ is the low pass filter and $\psi$ is the wavelet in work [1]. For the dual-tree complex wavelet, we let $\phi = (1/2\sqrt{2})\phi_1$ and $\psi = (1/8)\psi_1$ satisfying equation (10) with $\epsilon = 0.99$. $\phi_1$ and $\psi_1$ are defined in work [42, 43]. The fractional scattering propagator $\hat{W}x(u) = \{x \Theta_\alpha \phi_{2^J}(u), |x \Theta_\alpha \psi_\lambda(u)|\}_{\lambda \in P}$ is obtained with a

fractional wavelet transform $W$ followed by a modulus, which are both non-expansive. So $\hat{W}$ is also non-expansive. Since $S$ iteratively applies $\hat{W}$, it is also non-expansive. If $W$ is unitary, then $\hat{W}$ also preserves the signal norm $\|\hat{W}x\|^2 = \|x\|^2$. The FrScatNet is built layer by layer by iterating on fractional scattering propagator $\hat{W}$. If $\hat{W}$ preserves the signal norm, then the signal energy is equal to the sum of the fractional scattering energy of each layer plus the energy of the last propagated layer [1, 18]:

$$\|x\|^2 = \sum_{m=0}^{\bar{m}} \sum_{p \in P^m} \|S[p]x\|^2 + \sum_{p \in P^{\bar{m}+1}} \|U[p]\|^2 \quad (10)$$

By letting the network depth $\bar{m}$ tend to infinity, it results that the FrScatNet preserves the signal energy [1, 18]:

$$\|x\|^2 = \sum_{p \in P_\infty} \|S[p]x\|^2 = \|Sx\|^2 \quad (11)$$

The output of the FrScatNet is obtained by cascading the fractional scattering coefficients of every layer. The fractional windowed scattering $S$ is computed with a cascade of wavelet modulus operators, thus its properties depend upon the FRWT properties. The fractional scattering process is implemented along the frequency decreasing paths where most of the fractional scattering energy is concentrated. A frequency decreasing path $p = (2^{-j_1}r_1, ..., 2^{-j_m}r_m)$ satisfies $0 < j_k \leq j_{k+1} \leq J$. The fractional scattering coefficients are the lower frequencies filtered by the low pass filter, due to the energy of the signal mainly distributed on the lower frequency region. The signal analysis capability of the classical WT is limited in the time-frequency plane. If the distribution of the signal energy is not most concentrated on the frequency domain, the feature vector extracted by the network is not the best for the signal discrimination. Therefore, we extend the classical WT to the FRWT, which is able to offer signal energy representations in the time-fractional-frequency plane, and try to find the appropriate fractional order for the signal representation.

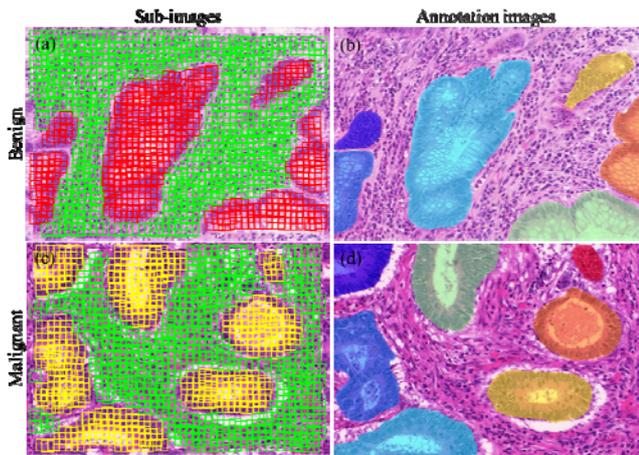

Fig.3. Sub-images obtained by the proposed method. Green rectangles represent the sub-images from the background regions. Red rectangles denote the sub-images from the benign areas. Yellow rectangles denote the sub-images from the malignant areas.

Image textures can be modeled as realization of stationary processes. The spectrum method is an important tool for texture analysis. The power spectrum only depending on the second-order moments is then not sufficient to discriminate image textures whose second-order moments are very similar. The scattering representation of the stationary processes depends upon second-order and normalized higher order moments. Thus, it can discriminate textures having the same second-order moments but different higher order moments [1]. In the proposed method, the fractional windowed scattering is computed based on the fractional wavelet transform, which can represent the image energy distribution in the fractional wavelet scattering domain. Therefore, the scattering implementation can be achieved through the path where most of the signal energy is concentrated in the fractional wavelet scattering domain, and the discrimination capability of the first order fractional scattering coefficients is improved. We can see that the first-order scattering coefficients depicted in Fig. 1(b), estimated from each realization, are nearly the same in the ScatNet while their second-order scattering coefficients shown in Fig. 1(c) are different. However, in the FrScatNet both the first- and second- order fractional scattering coefficients displayed in Fig. 1(d, e) are clearly different and have the discriminatory ability.

*C. Classification*

A scattering transform eliminates the image variability due to translations and linearizes small deformations [1]. Classification is then carried out with a Gaussian kernel SVM and a generative PCA classifier, respectively. The overall framework of the FrScatNet is illustrated in Fig. 2.

Firstly, a three-layer FrScatNet is constructed based on fractional wavelet transform. Secondly, all the normalized data are put into the FrScatNet. The characteristic matrix **Q**, of size of $L \times N \times D$, is obtained by concatenating the fractional scattering coefficients of all three layers. $L$ is the length of the feature vector detected from each input signal. $N$ denotes the number of the total input signals. $D$ represents the number of the fractional orders. In the next step, the input data are randomly divided into training and testing sets. Therefore, the corresponding characteristic matrix **Q** is also divided into the training characteristic matrix $\mathbf{Q}_{Train}$, the testing characteristic matrix $\mathbf{Q}_{Test}$, and termed as $\mathbf{Q} = [\mathbf{Q}_{Train}\ \mathbf{Q}_{Test}]$.

## III. EXPERIMENTS AND RESULTS

We test here the classification performance of the FrScatNet on two-dimensional signals and analyze its properties. We also propose an algorithm to achieve gland segmentation from the histology images.

*A. Image Classification, Application and Discussion*

In the proposed framework, we generalize the Morlet wavelet and dual-tree complex wavelet to fractional wavelets according to equation (3). These two corresponding fractional wavelets are used to construct the FrScatNet and compute the fractional scattering representations, respectively. So two sets of characteristic matrix are created to test the performance of the proposed framework and analyze the signal energy

TABLE I
CLASSIFICATION ERRORS IN TERMS OF FRACTIONAL WAVELET SCATTERING NETWORK IN VARIOUS ONE-DIMENSIONAL DATABASES.

| Dt | Fw | Cf | Rt | Fractional order($\alpha_1=1, \alpha_2$) | | | | | | | | | Fractional order($\alpha_2=1, \alpha_1$) | | | | | | | | |
|---|---|---|---|---|---|---|---|---|---|---|---|---|---|---|---|---|---|---|---|---|---|
| | | | | 0.05 | 0.10 | 0.40 | 0.70 | 1.00 | 1.30 | 1.60 | 1.90 | 1.95 | 0.05 | 0.10 | 0.40 | 0.70 | 1.00 | 1.30 | 1.60 | 1.90 | 1.95 |
| BD | I | PCA | 0.5 | 0.218 | 0.241 | **0.108** | 0.187 | 0.204 | 0.196 | 0.185 | 0.231 | 0.226 | 0.235 | 0.264 | 0.197 | **0.117** | 0.204 | 0.200 | 0.200 | 0.251 | 0.251 |
| | | | 0.8 | 0.241 | 0.274 | **0.113** | 0.192 | 0.220 | 0.217 | 0.203 | 0.254 | 0.248 | 0.264 | 0.289 | 0.210 | **0.121** | 0.220 | 0.217 | 0.216 | 0.272 | 0.269 |
| | | SVM | 0.5 | 0.213 | 0.257 | **0.156** | 0.200 | 0.260 | 0.202 | 0.200 | 0.249 | 0.199 | 0.228 | 0.251 | 0.243 | 0.227 | 0.260 | 0.259 | **0.155** | 0.250 | 0.210 |
| | | | 0.8 | 0.239 | 0.276 | **0.206** | 0.219 | 0.292 | 0.229 | 0.223 | 0.295 | 0.218 | 0.269 | 0.285 | 0.267 | 0.261 | 0.292 | 0.277 | **0.201** | 0.273 | 0.259 |
| | II | PCA | 0.5 | 0.243 | 0.274 | **0.129** | 0.237 | 0.303 | 0.234 | 0.227 | 0.296 | 0.242 | 0.284 | 0.282 | 0.277 | 0.277 | 0.303 | 0.284 | 0.259 | 0.291 | **0.133** |
| | | | 0.8 | 0.251 | 0.303 | **0.147** | 0.228 | 0.345 | 0.243 | 0.241 | 0.295 | 0.251 | 0.292 | 0.293 | 0.287 | 0.292 | 0.345 | 0.276 | 0.277 | 0.308 | **0.152** |
| | | SVM | 0.5 | 0.227 | 0.263 | 0.216 | 0.228 | 0.302 | 0.230 | **0.185** | 0.267 | 0.233 | 0.257 | 0.272 | 0.244 | 0.258 | 0.302 | 0.246 | **0.173** | 0.328 | 0.230 |
| | | | 0.8 | 0.228 | 0.276 | 0.219 | 0.228 | 0.312 | 0.232 | **0.218** | 0.270 | 0.233 | 0.264 | 0.257 | 0.254 | 0.275 | 0.312 | 0.250 | **0.202** | 0.331 | 0.242 |
| MD | I | PCA | 0.5 | 0.241 | 0.237 | **0.141** | 0.232 | 0.148 | 0.239 | 0.221 | 0.227 | 0.233 | 0.241 | 0.255 | 0.220 | 0.237 | **0.148** | 0.240 | 0.229 | 0.246 | 0.230 |
| | | | 0.8 | 0.283 | 0.272 | **0.115** | 0.277 | 0.138 | 0.279 | 0.262 | 0.302 | 0.258 | 0.262 | 0.286 | 0.265 | 0.289 | **0.138** | 0.288 | 0.269 | 0.302 | 0.267 |
| | | SVM | 0.5 | 0.291 | 0.280 | 0.277 | 0.290 | **0.154** | 0.297 | 0.270 | 0.289 | 0.274 | 0.292 | 0.333 | 0.285 | 0.294 | **0.154** | 0.298 | 0.294 | 0.318 | 0.308 |
| | | | 0.8 | 0.310 | 0.314 | 0.320 | 0.335 | **0.183** | 0.325 | 0.315 | 0.329 | 0.305 | 0.319 | 0.356 | 0.334 | 0.333 | **0.188** | 0.355 | 0.325 | 0.370 | 0.346 |
| | II | PCA | 0.5 | 0.255 | 0.265 | 0.247 | 0.259 | 0.276 | **0.188** | 0.244 | 0.279 | 0.245 | 0.268 | 0.317 | 0.255 | 0.273 | 0.276 | 0.264 | **0.185** | 0.320 | 0.259 |
| | | | 0.8 | 0.289 | 0.298 | 0.270 | 0.289 | 0.333 | **0.204** | 0.280 | 0.290 | 0.292 | 0.302 | 0.33 | 0.294 | 0.301 | 0.333 | 0.299 | **0.208** | 0.364 | 0.271 |
| | | SVM | 0.5 | 0.289 | 0.322 | 0.295 | 0.285 | 0.336 | 0.293 | **0.172** | 0.371 | 0.286 | 0.288 | 0.326 | 0.300 | 0.304 | 0.336 | 0.309 | **0.175** | 0.386 | 0.289 |
| | | | 0.8 | 0.315 | 0.356 | 0.312 | 0.320 | 0.344 | 0.324 | **0.118** | 0.366 | 0.312 | 0.329 | 0.357 | 0.326 | 0.327 | 0.344 | 0.328 | **0.203** | 0.397 | 0.312 |
| TSD | I | PCA | 0.5 | 0.440 | 0.401 | 0.442 | 0.487 | **0.161** | 0.487 | 0.423 | 0.394 | 0.453 | 0.363 | 0.396 | 0.412 | 0.405 | **0.161** | 0.408 | 0.398 | 0.411 | 0.363 |
| | | | 0.8 | 0.353 | 0.369 | 0.418 | 0.460 | **0.086** | 0.445 | 0.382 | 0.331 | 0.359 | 0.282 | 0.305 | 0.347 | 0.338 | **0.086** | 0.344 | 0.336 | 0.346 | 0.275 |
| | | SVM | 0.5 | 0.415 | 0.401 | 0.405 | 0.458 | **0.190** | 0.453 | 0.409 | 0.389 | 0.415 | 0.367 | 0.394 | 0.377 | 0.423 | **0.190** | 0.413 | 0.389 | 0.381 | 0.363 |
| | | | 0.8 | 0.333 | 0.332 | 0.333 | 0.380 | **0.120** | 0.385 | 0.365 | 0.310 | 0.332 | 0.312 | 0.34 | 0.348 | 0.352 | **0.120** | 0.333 | 0.327 | 0.340 | 0.303 |
| | II | PCA | 0.5 | 0.545 | 0.498 | 0.544 | 0.600 | **0.281** | 0.577 | 0.526 | 0.483 | 0.542 | 0.498 | 0.548 | 0.518 | 0.562 | **0.281** | 0.549 | 0.545 | 0.506 | 0.528 |
| | | | 0.8 | 0.501 | 0.455 | 0.497 | 0.551 | **0.178** | 0.548 | 0.492 | 0.46 | 0.488 | 0.406 | 0.502 | 0.517 | 0.495 | **0.178** | 0.505 | 0.505 | 0.426 | 0.457 |
| | | SVM | 0.5 | 0.475 | 0.435 | 0.461 | 0.511 | **0.276** | 0.525 | 0.453 | 0.443 | 0.460 | 0.451 | 0.495 | 0.432 | 0.491 | **0.276** | 0.490 | 0.456 | 0.479 | 0.483 |
| | | | 0.8 | 0.432 | 0.400 | 0.410 | 0.480 | **0.133** | 0.447 | 0.402 | 0.380 | 0.418 | 0.400 | 0.413 | 0.405 | 0.445 | **0.133** | 0.423 | 0.450 | 0.427 | 0.397 |

Dt: Database; Fw: Framework; Cf: Classifier; Rt: Ratio.

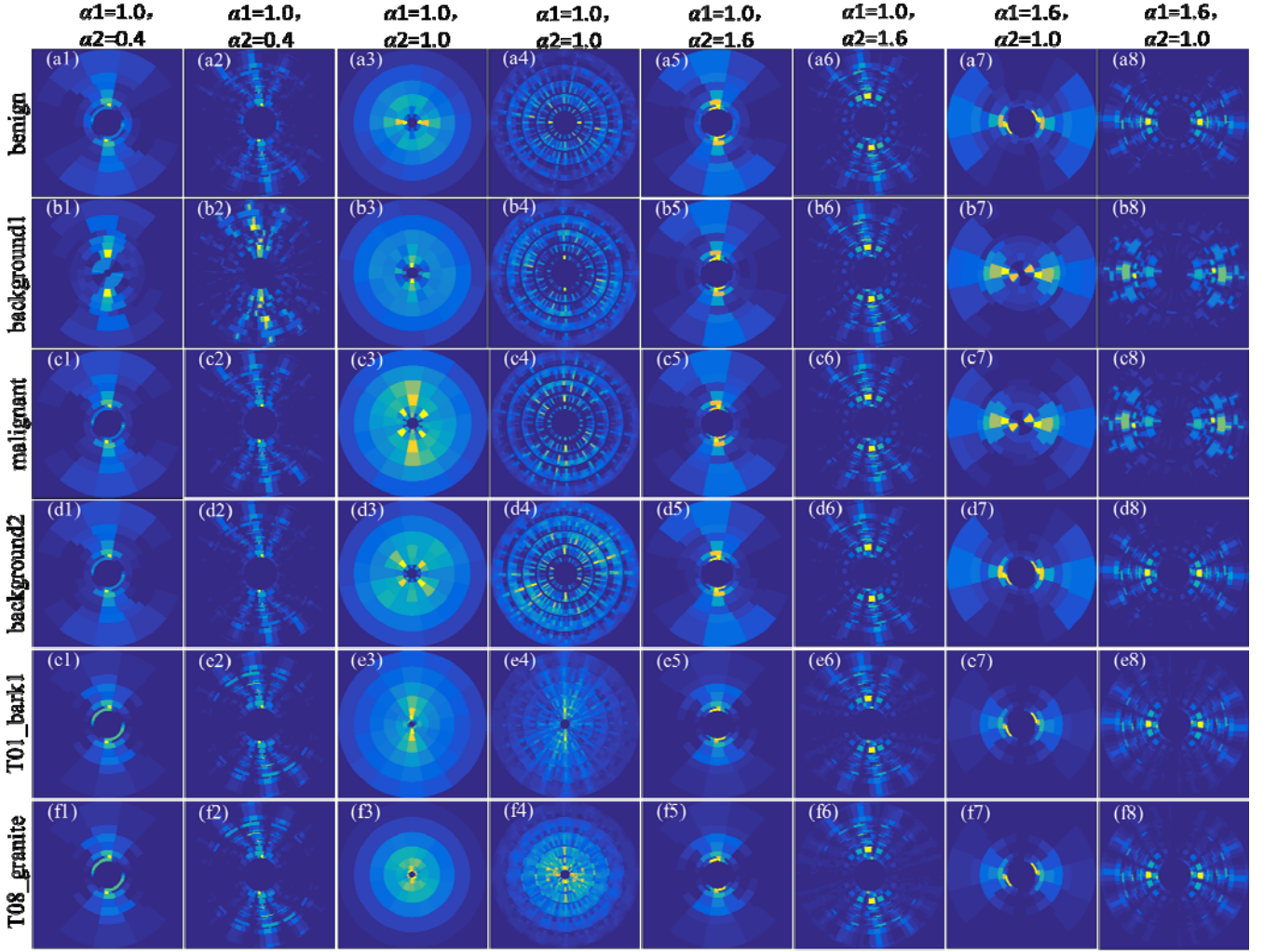

Fig.4. Fractional scattering coefficients of two-dimensional images from the FrameworkI in different fractional orders. (a1-a8) are of a random benign image from the DB. (b1-b8) are of a random background image from the DB. (c1-c8) are from a random malignant image from the MB. (d1-d8) are from a random background image from the MB. (e1-e8) are computed from a random image in T01_bark1 of TSD. (f1-f8) are computed from a random image in T08_granite of TSD; (a1-f1) are first-order fractional scattering coefficients when $\alpha_1=1.0$ and $\alpha_2=0.4$. (a2-f2) are second-order fractional scattering coefficients when $\alpha_1=1.0$ and $\alpha_2=0.4$. (a3-f3) are first-order scattering coefficients when $\alpha_1=1.0$ and $\alpha_2=1.0$. (a4-f4) are second-order scattering coefficients when $\alpha_1=1.0$ and $\alpha_2=1.0$. (a5-f5) are first-order fractional scattering coefficients when $\alpha_1=1.0$ and $\alpha_2=1.6$. (a5-f5) are second-order fractional scattering coefficients when $\alpha_1=1.0$ and $\alpha_2=1.6$. (a7-f7) are first-order fractional scattering coefficients when $\alpha_1=1.6$ and $\alpha_2=1.0$. (a8-f8) are second-order fractional scattering coefficients when $\alpha_1=1.6$ and $\alpha_2=1.0$

distributions. The FrScatNet constructed by the fractional complex Morlet wavelets identified as **FrameworkI** is initialized with the finest scale $2^J = 16$ and the total number of angles $K = 8$. The band-pass filters are constructed using the Gabor wavelets and the low frequency band is a Gaussian. The 2D fractional dual-tree complex wavelet based scattering network denoted as **FrameworkII** is initialized with five stages and six distinct orientations. In each direction, one of the two wavelets can be interpreted as the real part of a 2D complex-valued wavelet, the other as its imaginary part [42, 43]. The fractional orders for both frameworks are the same. For the two-dimensional wavelet, two fractional orders $\alpha_1$ and $\alpha_2$ are needed to determine the rotational angle. The angle is defined as $\theta = \alpha\pi/2$ ranging from 0 to $\pi$, so the fractional order $\alpha$ changes from 0 to 2. To save computation time, we fix one order as 1.00 and the other one changes within the range zero to two for computing the fractional scattering coefficients. The chosen values are 0.05, 0.10, 0.40, 0.70, 1.00, 1.30, 1.60, 1.90, and 1.95, respectively. The corresponding FRWT reduces to conventional WT when $\alpha_1 = \alpha_2 = 1.00$. Their performance is evaluated by comparing the classifier outputs.

The feature matrix obtained from FrameworkI is then put into two different kinds of classifiers named as PCA and SVM to achieve the classification, respectively. So does the feature matrix from FrameworkII. The dimensions of principal component are chosen as 10, 15, 20, 25, 30, 35, 40, 45, 50, 60, 70, and 80 in the PCA classifier. For the SVM classifier, the best-performing set of parameters is identified by calculating the five-fold cross-validation error over the trainings for each combination of parameters. The parameter C is set to $2^0$, $2^4$, $2^8$ and the parameter gamma is set to $2^{-16}$, $2^{-12}$, $2^{-8}$. The input data are divided into training and testing sets randomly. We set the ratio as 1:1 and 4:1 in our experiment. The classification results are averaged over 5 different random splits for each ratio.

*1) Warwick-QU Dataset*

The training set Warwick-QU includes 85 original pathological color images and their corresponding annotated binary images. The annotation is shown in Fig. 3 (b) and (d).

TABLE II
CLASSIFICATION ERRORS IN TERMS OF FRACTIONAL WAVELET SCATTERING NETWORK IN H&E TRAINING DATABASES

| Data | Ratio | Fractional order($\alpha_1$, $\alpha_2$) | | | | | | | | | | | | | | | |
|---|---|---|---|---|---|---|---|---|---|---|---|---|---|---|---|---|---|
| | | (0.4, 0.4) | (0.7, 0.4) | (1.3, 0.4) | (1.6, 0.4) | (0.4, 0.7) | (0.7, 0.7) | (1.3, 0.7) | (1.6, 0.7) | (0.4, 1.3) | (0.7, 1.3) | (1.3, 1.3) | (1.6, 1.3) | (0.4, 1.6) | (0.7, 1.6) | (1.3, 1.6) | (1.6, 1.6) |
| BD | 0.5 | 0.249 | 0.248 | 0.228 | 0.244 | 0.236 | 0.247 | 0.243 | 0.250 | **0.208** | 0.238 | 0.227 | 0.245 | 0.224 | 0.232 | 0.255 | 0.278 |
| | 0.8 | 0.231 | 0.239 | **0.218** | 0.237 | 0.249 | 0.245 | 0.236 | 0.232 | 0.219 | 0.229 | 0.235 | 0.276 | 0.252 | 0.247 | 0.248 | 0.267 |
| MD | 0.5 | **0.229** | 0.239 | 0.284 | 0.263 | 0.280 | 0.245 | 0.239 | 0.310 | 0.292 | 0.253 | 0.286 | 0.297 | 0.288 | 0.296 | 0.304 | 0.292 |
| | 0.8 | 0.259 | 0.261 | 0.264 | 0.255 | 0.269 | 0.242 | **0.234** | 0.291 | 0.271 | 0.249 | 0.276 | 0.293 | 0.282 | 0.286 | 0.305 | 0.284 |

The training set is combined by 37 color images with gland type labelled as "Benign" and 48 color images with gland type identified as "Malignant". Two sets of databases are generated from the training set to test the properties of the proposed framework.

We construct the benign and background database (BD) from the 37 original pathological color images with benign glands. The corresponding annotated image is used as a mask to determine the benign gland area. Firstly, a k-mean based algorithm [44] aggregates nearby pixels into super-pixels of nearly uniform size for each original images. Then, the original image with size of $m \times n \times 3$ is enlarged to obtain a new image with size of $(m+16) \times (n+16) \times 3$ by a mirror method. In the next, a sliding block window with size of $32 \times 32 \times 3$ is positioned at the center of these super-pixels. If the overlapping area between the sliding window and the irregular benign area is more than 95%, then the obtained sub-image is classified as belonging to benign database, or it is denoted as background database. In this way, we obtain a total of 7631 sub-images shown as the red rectangles from the benign areas and 12529 sub-images as the green rectangles from the background areas of 37 original pathological color images with benign gland in Fig. 3 (a). The same process is also applied to each of the original images issued from the 48 original pathological color images with malignant glands to construct the malignant and background database (MD). There are 15857 sub-images illustrated as the yellow rectangles from the malignant areas and 10782 sub-images shown as the green rectangles from the background areas in Fig. 3 (c). These two constructed sets are used for training.

All normalized sub-images from the databases are put into the FrScatNet. For the BD, the obtained feature matrix **Q** corresponding to FrameworkI and FrameworkII have respectively the size $681 \times 20160 \times 18$ and $391 \times 20160 \times 18$. For the MD, the obtained **Q** in the same situation corresponds respectively to $681 \times 26639 \times 18$ and $391 \times 26639 \times 18$. The classification result is shown in TABLE I. For the BD, the classification error is minimum in the FrameworkI when the fractional orders are set to $\alpha_1 = 1.00$ and $\alpha_2 = 0.40$ by using the PCA classifier and the fractional orders $\alpha_1 = 1.60$ and $\alpha_2 = 1.00$ for the SVM classifier. In the FrameworkII, the minimum classification error happens when $\alpha_1 = 1.00$ and $\alpha_2 = 0.40$ with the PCA classifier and, with SVM, $\alpha_1 = 1.60$ and $\alpha_2 = 1.00$. We can see that the classification result is better in the fractional wavelet scattering domain than the wavelet scattering domain for the BD, and the performance from FrameworkI is better than that from the FrameworkII. The fractional scattering coefficients of images from the benign group (Fig. 4. a1-a8) and the background group (Fig. 4. b1-b8) of BD detected by the FrameworkI are shown in Fig. 4. The first- and second-order fractional scattering coefficients (Fig. 4. a1-a2, b1-b2, a7-a8, b7-b8) from the FrScatNet are clearly different.

For the MD, in the FrameworkI, the error is minimum when $\alpha_1 = 1.00$ and $\alpha_2 = 0.40$ when using PCA classifier and $\alpha_1 = 1.00$ and $\alpha_2 = 1.00$ (the classical wavelet transform) when applying SVM. In the FrameworkII, the best classification occurs when $\alpha_1 = 1.60$ and $\alpha_2 = 1.00$ with PCA and $\alpha_1 = 1.00$ and $\alpha_2 = 1.60$ with SVM. We get the conclusion that for most MD cases, the best classification is achieved in the fractional wavelet scattering domain. The fractional scattering coefficients of images belonging to the MD resulting from FrameworkI are also displayed in Fig. 4. c1-c8 and Fig. 4. d1-d8. The first- and second-order scattering coefficients (Fig. 4. c3-c4, d3-d4) from the classical ScatNet are clearly different, and those fractional scattering coefficients (Fig. 4. c7-c8, d7-d8) from the FrScatNet are also clearly different.

*2) Textured Surfaces Dataset*

All normalized images from the texture surfaces dataset (TSD) are input into the FrScatNet. The size of matrix **Q** is respectively $681 \times 1000 \times 18$ and $391 \times 1000 \times 18$ for Frameworks I and II. The classification results are depicted in TABLE I. The best classification performance is obtained when $\alpha_1 = 1.00$ and $\alpha_2 = 1.00$, which means that the classification is better in the wavelet scattering domain than the fractional wavelet scattering domain. We display the scattering coefficients computed by the FrameworkI in Fig. 4 (e1-e8, f1-f8). By comparing the coefficients computed from the FrScatNet, it clearly appears that the first and second order scattering coefficients of ScatNet are different. Therefore, the classification performance is better in the wavelet scattering domain.

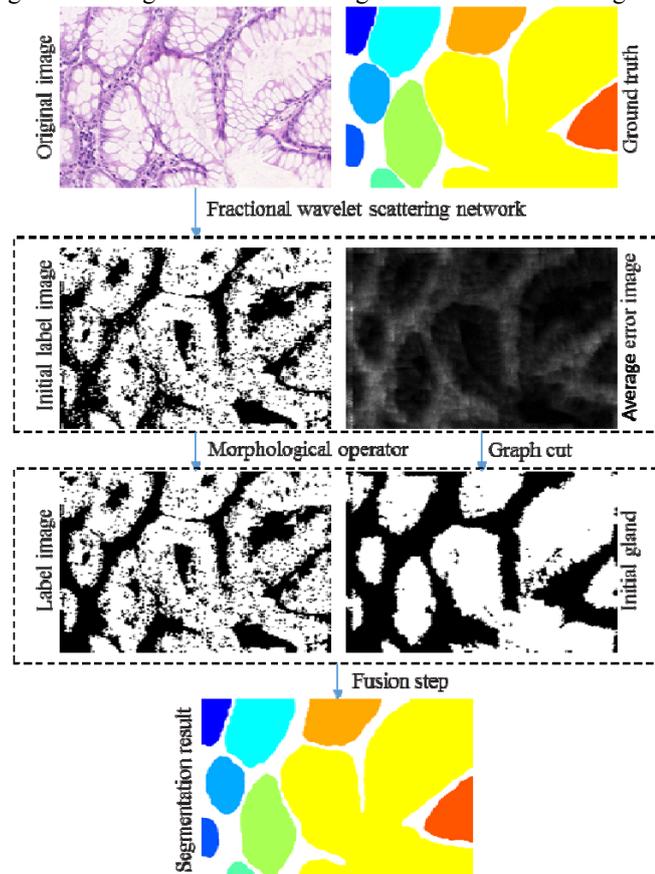

Fig.5. The framework structure of the proposed gland segmentation algorithm.

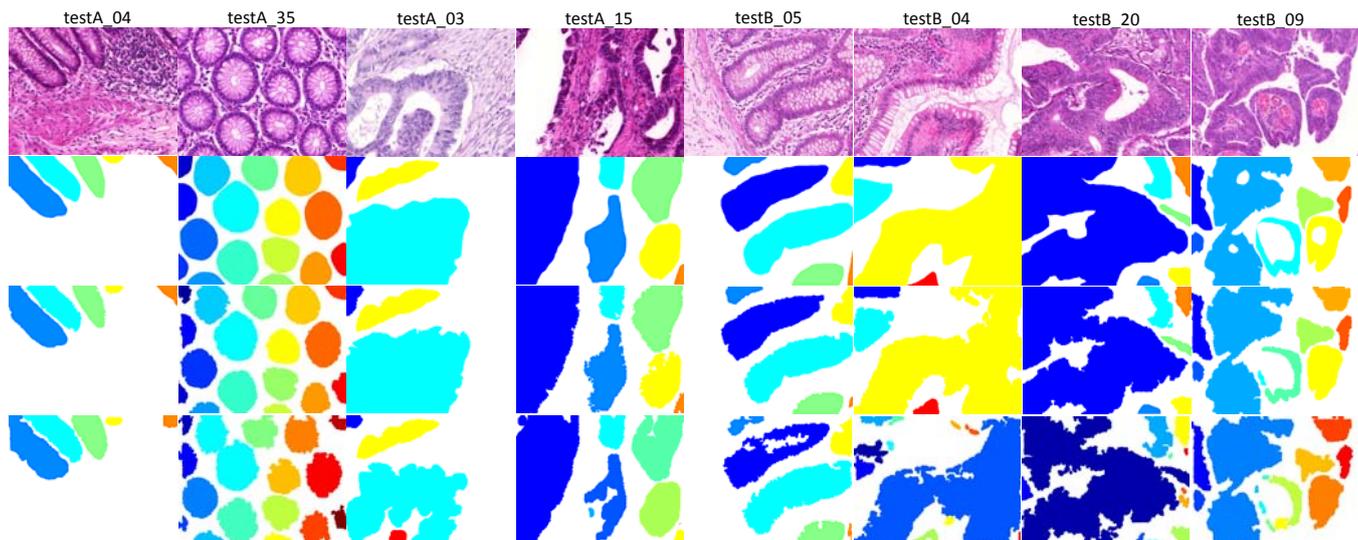

Fig.6. Segmentation results of test cases (from left to right): benign, benign, malignant, malignant, benign, benign, malignant, malignant cases. (from top to bottom): original images, ground truth, segmentation results by FrScatNet, and segmentation results by ScatNet (different colors individual gland objects).

In conclusion, the classification errors are minimum when $\alpha_1 = 1.00$, $\alpha_2 = 0.40$ for BD and MD, and $\alpha_1 = 1.00$, $\alpha_2 = 1.00$ for TSD. This shows that the proposed FrScatNet based on different FRWT is able to represent images in the fractional wavelet scattering domain and to provide the best fractional order for classification.

*B. Gland Segmentation in Colon Histology Images*

We test the FrScatNet on the dataset provided in the MICCAI challenge contest [35] aiming at gland segmentation. The rules defining the training set and the test group are followed as specified in the challenge.

*1) Feature Detection*

The test group is divided into Test A and Test B as indicated in section II. A. Test A contains 33 histological images with benign glands and 27 images with malignant glands. Test B includes 4 benign images and 16 malignant images. We estimate the best parameter on the training subset. According to the experiments shown in TABLE I, the fractional orders (including 0.4 0.7 1.3 1.6) appear as promising for classification. Here, the obtained fractional orders are tested by cross validation to determine the best classification parameter for the H&E images. According to the results shown in TABLE I, we select FrameworkI and PCA classifier. The validation results are shown in TABLE II. By comparing the classification results in TABLE II and I, we see that the best classification is achieved when $\alpha_1 = 1.00$, $\alpha_2 = 0.40$, Ratio = 0.50 for the benign images and $\alpha_1 = 1.00$, $\alpha_2 = 0.40$, Ratio = 0.80 for the malignant images, which are used for the following gland segmentation.

For each of the test group images, we try to classify it into two classes: the target (benign or malignant) region and the background region. To achieve this goal, we firstly divide the original pathological color images into sub-images of the same size 32×32×3 in the training data. We apply a k-mean based algorithm [44] to aggregate nearby pixels into super-pixels of nearly uniform size. Their boundaries closely match true target boundaries. Then, the original image is divided into sub-images as described above. Each sub-image is entered into the FrScatNet to get the feature matrix in the fractional wavelet scattering domain including wavelet scattering domain ($\alpha_1 = 1.00$, $\alpha_2 = 1.00$) as comparing group. The PCA classifier allows determining the class of super-pixels (the background is labeled as 1 and the target is labeled as 2) and the corresponding average approximation error for each instance and class pair. We get a fully labeled color image and the average approximation error for each class from this step.

*2) Gland Segmentation*

Some initially labelled images have misclassified areas within the glands and need to be corrected. In a first step, we use an erosion operator to reduce these misclassifications and one dilation step to keep the gland shape. The obtained result is regarded as the labelled image shown in Fig. 5. The gland boundaries are better delineated in this new labelled image but some areas of the central regions are lost mainly. This happens because some regions within the gland structures have similar features to the background region. But the complex gland frames are detected due to the fact that FrScatNet is mainly based on the texture information but not shape feature. The epithelial cells are lined up around a lumen to form glandular structures. Although the cancer leads to morphological and structural deviations, the texture information especially within the boundary regions is stable and effective. Besides, very close or even touching glands make also the segmentation difficult. The average error image is used to handle these problems in this study. We use the graph cut method [45] to classify the average approximation error into two classes: the foreground region and the background region. The initial glands shown in Fig. 5 are used to locate the gland structures in the image. They are usually smaller than the corresponding annotated structure. A region growing algorithm, regarded as the fusion step in Fig. 5, is applied to increase the gland area in order to better fit the true gland regions and to keep touching glands separated. Then, a simple erosion operation is performed to delete isolated points followed by a dilation operation for preserving the gland shapes.

*3) Evaluation*

To visually evaluate the performance of the method, some segmentation results on the testing dataset are shown in Fig. 6. A comparison between FrScatNet and the ScatNet is also provided. As it can be seen in the third and the fourth rows, FrScatNet leads to a better segmentation in both benign and malignant cases from test A and test B. In the ScatNet based method, some touching glands cannot be separated while they are with FrScatNet. Besides, some over- or under-segmentation problems can be observed. This is not surprising because the fractional transform makes it possible to separate different signals by successive rotations and filtering on the indicated side of the axes using some low or high-pass filter [46]. In addition, the extra parameter obtained by rotating angles in the fractional transform gives an additional degree of freedom that can be used to optimize the performance of the network. The correlation peak within a certain class of signals can be made more prominent or sharper in an optimally chosen fractional scattering domain [46, 47]. Last, the traditional wavelet transform detects too many edge details in histology image leading to unsmoothed boundaries and deviations from the ground truth while the fractional wavelet transform offers a better balance between these two issues.

However, these latter problems still remain when applying our method. Non-smoothed boundaries are mainly caused by the construction of the super-pixel regions. It is also challenging to recognize small background areas within a given target region as shown in Fig. 6.testB_09. It happens that the windows used to detect the feature vectors contain several target regions and the graph cut method classifies those little regions as target regions.

To quantitatively assess the performance of our method, we conduct comparisons with Fully Convolutional Network (FCN) [48], dilated FCN (DFCN) [49], Deep Multichannel Neural Networks (DMNN) [22], and some of the solutions proposed in the MICCAI challenge such as CUMedVision1 (CUM1), CUMedVision2 (CUM2) [34], Frerburg1 (Fre1), Frerburg2 (Fre2) [36], ExB1, ExB2, ExB3 , CVIP Dundee (Dund) and LIB [35]. All of them have used the datasets provided by the MICCAI 2015 Gland Segmentation Challenge Contest. The ground truth being known, the evaluation methods proposed in this competition include the accuracy of the detection of individual glands, the volume-based accuracy of the segmentation of individual glands and the boundary-based similarity between glands and their corresponding segmentation [35, 37]. A metric for the gland detection is the F1-score, defined by

$$F_{1score} = \frac{2 \cdot P \cdot R}{P+R}, \quad P = \frac{TP}{TP+FP}, \quad R = \frac{TP}{TP+FN}, \quad (12)$$

where *TP* is the number of true positives, *FP* is the number of false positives, and *FN* is the number of false negatives. The ground truth for each segmented glandular object is the glandular object in the manual annotation that shares maximum overlap with the segmented glandular object. A segmented object is considered as a true positive if it has more than 50% area overlap with its ground truth, or it is considered as a false positive. An object of the ground truth is regarded as a false negative if it has no corresponding prediction or has less than 50% area overlap with its predicted glandular object.

Two sets of pixels are defined as *G* representing the ground truth and *S* denoting the segmented objects. The Dice index is used for the evaluation of the segmentation on the whole image $D(G, S) = 2(|G \cap S|)/(|G|+|S|)$. An object-level Dice index can also be defined as:

$$D_{object}(G, S) = \frac{1}{2}\left[\sum_{i=1}^{n_S} \omega_i D(G_i, S_i) + \sum_{i=1}^{n_G} \hat{\omega}_i D(\hat{G}_i, \hat{S}_i)\right], \quad (13)$$

$$\omega_i = |S_i| / \sum_{j=1}^{n_S}|S_j|, \quad \hat{\omega}_i = |\hat{G}_i| / \sum_{j=1}^{n_G}|\hat{G}_j|. \quad (14)$$

where $n_G$ denotes the total number of ground truth objects in image and $n_S$ the total number of segmented objects in an image.

TABLE III
PERFORMANCE IN COMPARISON TO OTHER METHODS

| Method | $F_{1score}$ | | | | $D_{object}$ | | | | $H_{object}$ | | | | RS | WRS |
|---|---|---|---|---|---|---|---|---|---|---|---|---|---|---|
| | Part A | | Part B | | Part A | | Part B | | Part A | | Part B | | | |
| | Score | Rank | Score | Rank | Score | Rank | Score | Rank | Score | Rank | Score | Rank | | |
| CUM1 | 0.868 | 9 | 0.769 | 4 | 0.867 | 12 | 0.800 | 4 | 74.596 | 12 | 153.646 | 9 | 50 | 29 |
| CUM2 | **0.912** | 1 | 0.716 | 7 | 0.897 | 2 | 0.781 | 9 | 45.418 | 2 | 160.347 | 11 | 32 | 10.5 |
| ExB1 | 0.891 | 6 | 0.703 | 9 | 0.882 | 6 | 0.786 | 7 | 57.413 | 9 | 145.575 | 5 | 42 | 21 |
| ExB2 | 0.892 | 5 | 0.686 | 11 | 0.884 | 5 | 0.754 | 12 | 54.785 | 4 | 187.442 | 12 | 49 | 19.25 |
| ExB3 | 0.896 | 3 | 0.719 | 6 | 0.886 | 4 | 0.765 | 10 | 57.350 | 8 | 159.873 | 10 | 41 | 17.75 |
| Fre1 | 0.834 | 12 | 0.605 | 13 | 0.875 | 9 | 0.783 | 8 | 57.194 | 7 | 146.607 | 7 | 56 | 28 |
| Fre2 | 0.870 | 8 | 0.695 | 10 | 0.876 | 8 | 0.765 | 10 | 57.093 | 6 | 148.463 | 8 | 50 | 23.5 |
| Dund | 0.863 | 10 | 0.633 | 12 | 0.870 | 11 | 0.715 | 13 | 58.339 | 10 | 209.048 | 14 | 70 | 33 |
| LIB | 0.797 | 13 | 0.306 | 14 | 0.801 | 14 | 0.617 | 14 | 101.167 | 14 | 190.447 | 13 | 82 | 41 |
| FCN | 0.788 | 14 | 0.764 | 5 | 0.813 | 13 | 0.796 | 5 | 95.054 | 13 | 146.248 | 6 | 56 | 34 |
| DFCN | 0.854 | 11 | 0.798 | 3 | 0.879 | 7 | 0.825 | 3 | 62.216 | 11 | 118.734 | 3 | 38 | 24 |
| DMNN | 0.893 | 4 | 0.843 | 2 | **0.908** | 1 | 0.833 | 2 | **44.129** | 1 | **116.821** | 1 | 11 | 5.75 |
| **ScatNet** | 0.874 | 7 | 0.710 | 8 | 0.875 | 9 | 0.791 | 6 | 56.593 | 6 | 126.339 | 4 | 40 | 21 |
| **FrScatNet** | 0.901 | 2 | **0.858** | 1 | 0.896 | 3 | **0.842** | 1 | 52.276 | 3 | 117.100 | 2 | 12 | 7 |

$S_i$ is the $i$th segmented object in an image, and $G_i$ is the ground truth object that maximally overlaps $S_i$. $\hat{G}$ denotes the $i$th ground truth object in an image, and $\hat{S}$ denotes a segmented object that maximally overlaps $\hat{G}$ in the image.

The Hausdorff distance between $G$ and $S$ is often used to evaluate the shape similarity defined as

$$H(G,S) = \max\left\{\sup_{x \in G}\inf_{y \in S} f\|x-y\|, \sup_{y \in S}\inf_{x \in G} f\|x-y\|\right\}. \quad (15)$$

We can also measure the shape similarity between all segmented objects by using the object-level Hausdorff distance:

$$H_{object}(G,S) = \frac{1}{2}\left[\sum_{i=1}^{n_S}\omega_i H(G_i, S_i) + \sum_{i=1}^{n_G}\hat{\omega}_i D(\hat{G}_i, \hat{S}_i)\right], \quad (16)$$

$$\omega_i = |S_i|/\sum_{j=1}^{n_S}|S_j|, \quad \hat{\omega}_i = |\hat{G}_i|/\sum_{j=1}^{n_G}|\hat{G}_j|. \quad (17)$$

For the overall results, the final score is the summation of all rankings (RS) from test set A and test set B based on these three criteria. Smaller final ranking stands for a better segmentation performance. Another index, the weighted rank sum (WRS), is used based on the weighted average of three evaluation criteria on the two test sets and is defined as [22]:

$$WeightedRS = 3/4\sum testARank + 1/4\sum testBRank. \quad (18)$$

*4) Results and Discussion*

All the results are listed in TABLE III. For the gland detection evaluation, the results of the $F1$ score are shown in the left columns in TABLE III. The CUMedVision2 (CUM2) with a contour-aware component achieves the best results for test A but the performance of our method remains high. FrScatNet behaves better than all other methods on test B. For the gland segmentation evaluation, the results of the object-level Dice index ($D_{object}$ in TABLE III) show that the Deep Multichannel Neural Networks (DMNN) algorithm achieves the best performance on test A and our method on test B. For the shape similarity evaluation ($H_{object}$ column in TABLE III), the DMNN algorithm is also ranked first but the index value obtained with FrScatNet is very close.

The final ranking score (RS) and the weighted ranking score (WRS), combining region, location and edge information, show that the FrScatNet and the DMNN methods lead to almost similar results and perform well better than all other solutions. Several observations can be made on these results. If FrScatNet is able to well detect the gland texture features, to correctly locate the glands and to separate them when they touch each other, it may have difficulties in capturing their high variabilities. The deep convolution networks usually have learning filters to take into account these unknown variabilities [50]. The FrScatNet is only able to provide the first two layers of such network and eliminate translation or rotation variability.

But the FrScatNet is learning-free due to the fact that the wavelet bases are fixed. The second layer fractional scattering coefficients provide important complementary information with a small computational and memory cost. FrScatNet achieves a better performance than ScatNet, the fractional scattering coefficients being more stable and discriminative.

The performance of FrScatNet obtained on test B is encouraging because this set contains more cancerous glands with more complicated shapes and size variabilities. For malignant cases, cancer progresses may cause changes in the component organization, and also lead to tissue deviations from their normal appearances. Our method mainly utilizes texture information to detect the margin of glands and to overcome the unclear boundary problem. The DMNN method and also the CUMedVision1 and CUMedVision2 approaches integrate edge information and location context to improve the results on both test sets.

IV. CONCLUSION

In this paper, the FrScatNet has been proposed and implemented. It extends the traditional ScatNet to the fractional scattering domain and provides signal representation in the time-fractional-frequency plan to improve the signal classification and segmentation performance. An automated method for gland segmentation in histological images was also proposed based on the FrScatNet. The graph cut method was used to process the average approximation error image and to locate the glands. We compared the proposed method with those reported in the MICCAI 2015 Gland Segmentation Challenge. Experimental analysis showed stable and comparable results. Significantly better results for malignant objects were obtained. They could be further improved by integrating edge information. The proposed approach is flexible and generic enough to be considered for organ/tissue segmentation in multimodal medical image.

ACKNOWLEDGMENT

We thank the MICCAI 2015 Gland Segmentation Challenge for providing histology dataset and the Lazebnik's group for offering access to the texture database. We thank Mallat' group for offering access to the ScatNet software.